\documentclass[conference] {IEEEtran}
\usepackage{graphicx}
\usepackage{amsmath}
\usepackage{amsfonts}%
\usepackage{amssymb}
\usepackage{amsthm}  %
\usepackage{subfig}
\usepackage{tabularx}
\usepackage{multirow}
\usepackage{caption}
\title{\huge Multidomain Multimodal Fusion For Human Action Recognition Using Inertial Sensors}
\author{Zeeshan Ahmad \\
	Department of Electrical and Computer Engineering\\
	Ryerson University\\
	Toronto, Canada\\
	z1ahmad@ryerson.ca
	\and
	  Naimul Mefraz Khan\\
	  Department of Electrical and Computer Engineering\\
	  Ryerson University\\
	  Toronto, Canada\\
  n77khan@ee.ryerson.ca
}

\begin{document}
	\maketitle
	\begin{abstract}
		
		One of the major reasons for misclassification of multiplex actions during action recognition is the unavailability of complementary features that provide the semantic information about the actions. In different domains these features are present with different scales and intensities. In existing literature, features are extracted independently in different domains but the benefits from fusing these multidomain features are not realized. To address this challenge and to extract complete set of complementary information, in this paper, we propose a novel multidomain multimodal fusion framework that extracts complementary and distinct features from different domains of the input modality. We transform input inertial data into signal images, and then made the input modality multidomain and multimodal by transforming spatial domain information into frequency and time-spectrum domain using Discrete Fourier Transform (DFT) and Gabor wavelet transform (GWT) respectively. Features in different domains are extracted by Convolutional Neural networks (CNNs) and then fused by canonical correlation based fusion (CCF) for improving the accuracy of human action recognition. Experimental results on three inertial datasets show the superiority of the proposed method when compared to the state-of-the-art.

	\end{abstract} 
\begin{IEEEkeywords}
 Convolutional neural network, Human action recognition, Multimodal fusion.
	\end{IEEEkeywords}		
	\section{Introduction}
Human action recognition (HAR) is a dynamic field for the researchers working in the area of machine learning, computer vision and human computer interaction due to its increasing  applications in daily life such as healthcare, sports and security~\cite{aggarwal2011human}. Before the success of deep learning models, the traditional methods for HAR were dependent on hand crafted features where the feature extraction to classification was performed separately, and thus they did not leverage the advantages of end to end learning. \let\thefootnote\relax\footnote{© 2020 IEEE. Personal use of this material is permitted. Permission from IEEE must be obtained for all other uses, in any current or future media, including reprinting/republishing this material for advertising or promotional purposes, creating new collective works, for resale or redistribution to servers or lists, or reuse of any copyrighted component of this work in other works.}

With the availability of smart phones and inertial wearable sensors at reasonable cost, the HAR using inertail data has gained significant attention. Inertail sensors provide data in the from of multivariate time series, at high sampling rate and their working is independent on the availability of light and thus there are no problems of view point variations, light intensities and cluttering as in case of vision based HAR~\cite{yang2009distributed}. 

Deep neural networks, since their revival, has gained significant attention due to their crowd pulling performance in various machine learning and computer vision applications~\cite{krizhevsky2012imagenet}~\cite{hinton2012deep}.
 The capacity of deep learning models to learn complex features directly from the data is useful to recognize the less discriminative and complex human actions. Initially deep learning models for HAR were employed on single modalities and therefore they did not capture the complementary features necessary to recognize multiplex actions. To extract essential supportive information for classification of less distinct actions, many fusion frameworks have been proposed based on statistical and deep learning based methods~\cite{chen2015improving}~\cite{ahmad2018towards}. In these fusion frameworks, the complementary features are extracted only in spatial domain and thus they are uniscalar and monochromatic.

 To address this shortcoming, in this paper, we propose CNN based multidomain multimodal fusion framework that extract features in spatial, frequency and time-frequency domain and hence provide multiscale and multichromatic complementary features in addition to discriminative features. These multidomain features are integrated using canonical correlation based fusion to improve the accuracy of action recognition. The key contributions of the proposed work are:
 
 \begin{itemize}
 	
 	\item Transform unimodal input into multidomain and multimodal to extract discriminative and complementary features in different domains and then perform canonical correlation based fusion at two stages to improve the accuracy of recognition.
 	
 	\item We successfully convert inertial sensor data, which is in the form of multivariate time series, into signal images. Conversion of 1D temporal data into 2D information allows CNN to extract unique features that would not be possible with 1D temporal information.
 	
 \end{itemize}

	\begin{figure*}
	\centering
	\includegraphics[width=0.7\linewidth]{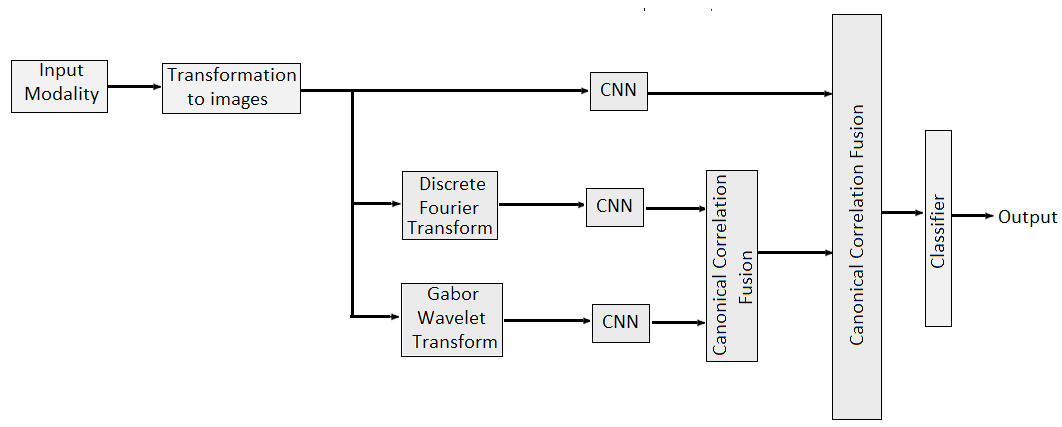}
	\caption{Complete Overview of Proposed Method}
	\label{fig: overview}
\end{figure*}

	\section{Related Work}
	
	Many methods have been proposed in literature for action recognition from inertial sensors. Here, we focus particularly on methods that utilize frequency domain and wavelet features, which serves as the motivation behind the multidomain features we are fusing. 
	
In~\cite{khare2017complex}, dual tree complex wavelet transform (DTCWT) is used to enhance the edge representation in the human action datasets which in turns improve the recognition accuracy of human actions.
	In~\cite{wang2013position} an improved algorithm based on Fast Fourier Transform (FFT) is proposed which extracts features from resultant acceleration of the data obtained from smartphones. Experiment shows that recognition accuracy is higher by selecting features in frequency domain.
	In~\cite{li2010multimodal}, a robust and accurate system was developed by performing multimodal fusion of ECG and accelerometer data in temporal and cepstral domains for physical activity recognition.
	In~\cite{yousefi2013biological}, biologically inspired motion analysis system is developed using different scales and orientations of Gabor wavelets to form a
	dictionary regarding human recognition. The proposed system shows promising performance on KTH human action dataset.	
	In~\cite{liu2019multi}, an effective multidomain and multi-task learning framework (MDMTL) is proposed to address the problems of domain-invariant feature learning and multi-task modeling for  human action recognition problem by performing fusion between view invariant and modality invariant features. 
	A fractional Fourier shape descriptor based on two-stage random forest based classification framework is proposed in~\cite{cai2015human} for silhouette based human  action recognition. The fractional Fourier shape representation of human
	silhouette is more powerful than that in the
	time or frequency domain. Diffusion score
	is used to determine best fractional order. 	
	In~\cite{shah2016encoding}, a frequency-based action descriptor called “FADE” is proposed to represent human actions. FFT is used to transform the signals into frequency domain and then these signals are resampled to exploit the frequency domain features. Finally Manhattan distance is used for measuring similarities between the actions. Experimental results show that the proposed method is computationally efficient.	
	In ~\cite{ulhaq2018space}, an advanced space-time
	filtering framework for recognizing human actions despite large
	viewpoint variations is proposed. Local motion in action sequence is characterized by 3D tensor structure for each pixel. Discrete Fourier transform is then applied to achieve frequency domain representation. Finally  view clusters are formed from multiple view
	action data and use space-time correlation filtering to achieve
	discriminative view representations to achieve action recognition.
	
	 In general, an information fusion analysis task involves processing of multi-modal data in order to obtain
	valuable insights about the data, a situation, or a higher level activity. Examples of information fusion
	analysis tasks include semantic concept detection, human action recognition, human tracking,
	event detection, etc. Multimedia data used for these tasks could be sensory (such as inertial, depth, RFID)
	or non-sensory (such as WWW resources, database). The fusion of multiple modalities can provide
	complementary information and increase the accuracy of the overall decision making process. 

	In all existing methods features are extracted independently in different domains, but fusion between different domains is not exploited. To address this challenge, in this paper, we propose a multidomain fusion framework that extract and fuse multidomain features simultaneously to get maximum benefits of multidomain fusion.


	\section{Proposed Method}
	
	In proposed method, we first transform input modality into images. Then, the input modality is converted to multidomain and multimodal by transforming spatial domain information into frequency and time-spectrum domain using Discrete Fourier transform (DFT) and Gabor wavelet transform, respectively. CNNs, whose architecture is shown in Fig.~\ref{fig:CNN Architecture for signal images}, are employed to extract multidomain features from the modalities and then canonical correlation based fusion is applied at two stages to get the final set of features. These features are finally fed to the SVM classifier for recognition task. The complete overview of the proposed method is shown in Fig.~\ref{fig: overview}. 
	
	\subsection{Formation of Signal Images}
	
\begin{figure}\label{signal images}
		\centering
		\includegraphics[width=\linewidth]{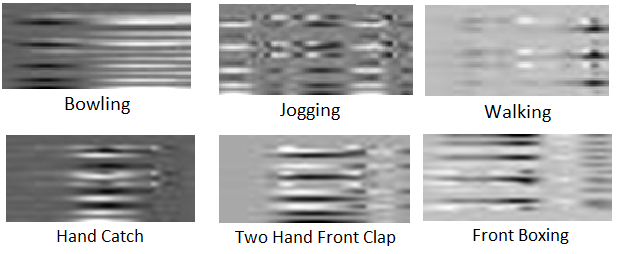}
		\caption{Signal Images of Six different actions}
		\label{fig:signal images}
	\end{figure}

	Before making input modality multidomain and multimodal, we transform input modality into signal images. The inertial sensors in our dataset are tri-axis accelerometer and gyroscopes, Thus, we have six sequences of signals. With the given six signal sequences,
	signal image is obtained through row-by-row stacking of raw
	signal sequences in such a way that each sequence comes
	in neighbor to every other sequence. Thus the width of our signal image becomes 24 according to the algorithm in~\cite{jiang2015human}. 
	 To decide on the length of the signal image, we made use of
	 sampling rate of datasets which is 50Hz.
	 Therefore, after certain experiments and to capture gritty motion accurately, the length of the signal image is finalized as 52, resulting in a final image
	 size of 24 x 52. The signal images are shown in Fig.~\ref{fig:signal images}.

	\subsection{ CNN Architecture}
	
\begin{figure*}
		\centering
		\includegraphics[width=\linewidth]{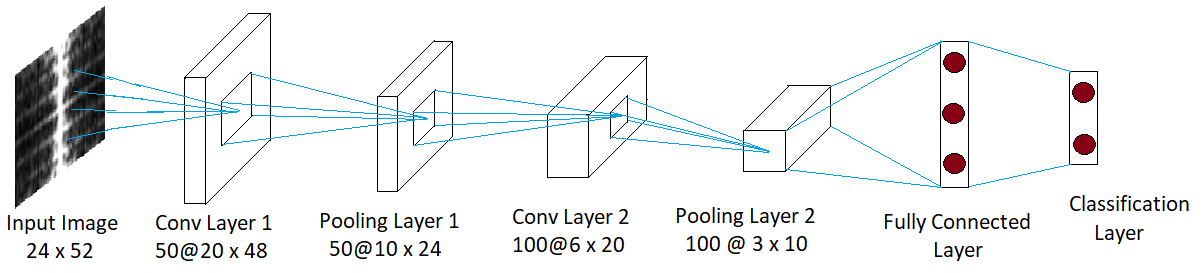}
		\caption{CNN Architecture for Signal Image}
		\label{fig:CNN Architecture for signal images}
		
	\end{figure*}

	The architecture
	of CNN used in proposed method is shown in Fig.~\ref{fig:CNN Architecture for signal images}. The architecture consists of an input layer, two convolutional layers, two pooling
	layers, a fully connected layer and a classification layer. The size of input layer is same as the size of signal image. The first convolutional layer has 50 kernels of
	size 5x5, followed by pooling layer of size 2x2 and stride
	2. The second convolutional layer which has 100 kernels followed
	by 2x2 pooling layer with stride 2.

	\subsection{2D Discrete Fourier Transform}
	
	Discrete Fourier Transform converts the spatial domain images into frequency domain where each point represents a particular frequency contained in the spatial domain image.
	
	\begin{equation}
	I(u,v) = \sum_{x=0}^{M-1}\sum_{y=0}^{N-1}{I(x,y)}e^{-j2\pi(\frac{ux}{M}+\frac{vy}{N})}
	\end{equation}
	
	Where $I(x,y)$ is the spatial domain image of size $M \times N$.      
	
	\subsection{Gabor Wavelet Transform}
	
	Gabor wavelet transform converts the images into time-spectrum domain. 2D Gabor filter used for transformation is bi-dimensional 
	Gaussian function centered at origin (0,0) with variance $\sigma$ modulated by
	a complex sinusoid with polar frequency $(F,\omega)$ and phase $P$ described in equation~\ref{Gabor}.
	
	\begin{equation} \label{Gabor}
	gab =Ke^{-\pi\sigma^2(x^2+y^2)}e^{(j2\pi F(xcos\omega+ysin\omega)+P)}
	\end{equation}
	
	Where $K$ is the magnitude of Gaussian envelope.
	
	\subsection{Canonical Correlation Analysis}\label{CCA Fusion}
	Canonical correlation analysis is effective and robust multivariate statistical method for finding the relationship between two sets of variables.
	
	Let $X\in\mathbb{R}^{p \times n}$ and $Y\in\mathbb{R}^{q \times n}$ represents the feature matrices from two modalities containing $n$ training samples. let $\Sigma_{xx}\in \mathbb{R}^{p \times p}$ and $\Sigma_{yy}\in \mathbb{R}^{q \times q}$ denote the within set covariance matrices of $X$ and $Y$ respectively and $\Sigma_{xy}\in \mathbb{R}^{p \times q}$ 
	denotes the between set covariance matrix for $X$ and $Y$ and $\Sigma_{yx}=\Sigma_{xy}^T$. The overall augmented covariance matrix of size $(p+q)\times(p+q)$ is given by
	
	
	\begin{equation}\label{first correlation equation}
	cov(X,Y) = 
	\begin{bmatrix}
	
	\Sigma_{xx} & \Sigma_{xy} \\
	\Sigma_{yx} & \Sigma_{yy}
	
	\end{bmatrix}
	\end{equation}
	
	The purpose of CCA is to find the linear combination $X^\prime=AX$ and $Y^\prime=BY$ such that the maximum pairwise correlation between the dataset could be achieved. Matrices $A$ and $B$ are called transformation matrices for $X$ and $Y$ respectively. The correlation between $X^\prime$ and $Y^\prime$ is given by  
	
	\begin{equation}
	corr(X^\prime,Y^\prime) = \frac{cov(X^\prime,Y^\prime)}{var(X^\prime).var(Y^\prime)}
	\end{equation}
	where $cov(X^\prime,Y^\prime)=A^T\Sigma_{xy}B$, $var(X^\prime)= A^T\Sigma_{xx}A$ and $var(Y^\prime)= B^T\Sigma_{yy}B$. $X^\prime$ and $Y^\prime$ are known as canonical variates.
	
	Lagrange's Optimization method is used to maximize the covariance between $X^\prime$ and $Y^\prime$ subject to the constraint that the variance of $X^\prime$ and variance of $Y^\prime$ is equal to unity.
	\[
	var(X^\prime)=var(Y^\prime)=1 
	\]
	
	The transformation matrices $A$ and $B$ are finally obtained by solving eigen value problem~\cite{krzanowski2000principles}.
	\begin{equation}\label{eigen vector equation}
	\begin{aligned}
	\Sigma_{xx}^{-1}\Sigma_{xy}\Sigma_{yy}^{-1}\Sigma_{yx}\bar{a} = \wedge^2\bar{a} \\ 
	\Sigma_{yy}^{-1}\Sigma_{yx}\Sigma_{xx}^{-1}\Sigma_{xy}\bar{b} = \wedge^2\bar{b}
	\end{aligned}
	\end{equation}
	
	where $\bar{a}$ and $\bar{b}$ are the eigen vectors and $\wedge^2$ is the diagonal matrix of eigen values or square of the canonical correlations. The number of nonzero eigen values in (5) satify the following condition.
	\[
	N = rank(\Sigma_{xy})\leq min(n,p,q)
	\]
	
	$A$ and $B$ are the transformation matrices consist of sorted eigen vectors.
	
	The covariance matrix for the transformed data can be written as   
	
	\[
	cov(X^\prime,Y^\prime)=
	\left[\begin{array}{cccc|cccc}
	1           & 0        & \dots  & 0       & \lambda_1& 0        & \dots  & 0 \\
	0           & 1        & \dots  &0        &         0& \lambda_2&\dots   &0  \\
	\vdots      &          & \ddots &         & \vdots   &          & \ddots & \\
	0           & 0        & \dots  & 1       &         0&         0& \dots  &\lambda_N \\
	\hline
	\lambda_1 & 0        & \dots  & 0       & 	1    & 0        & \dots  & 0\\  
	0 & \lambda_2&\dots   &0        &	0        & 1        & \dots  &0   \\  
	\vdots    &          & \ddots &         &\vdots    &          & \ddots &   \\  
	0 &         0& \dots  &\lambda_N&0         & 0        & \dots  & 1  
	\end{array}\right]
	\]
	The above matrix indicates that the canonical variates $X^\prime$ and $Y^\prime$  have nonzero correlation only on their corresponding indices. The identity matrices in the upper left and lower right corners show that the canonical variates are uncorrelated within each data set.
	
	On the transformed feature vector, fusion can be performed either by summation or concatenation. Since we use summation, therefore the discriminant features obtained after fusion can be mathematically written as
	\[
	Z =  X^\prime + Y^\prime = A^T X + B^T Y \]
	
	In Matrix form
	
	\begin{equation}
	Z= \begin{bmatrix}
	A \\
	B
	\end{bmatrix}^T \begin{bmatrix}X\\Y \end{bmatrix}
	\end{equation}
		\subsection{Support Vector Machine}
	Support Vector Machines (SVM) are supervised classifiers used in various machine learning, data mining and computer vision applications.~\cite{burges1998tutorial}.
	In simplest form, the score function for SVM is the mapping of the input vector to the scores and is a simple matrix operation as shown in Equation~\ref{eq:svm}.
	\begin{equation}\label{eq:svm}
	f=Wx + b
	\end{equation}
	Where $x$ is the input vector, $W$ is the weight determined by input vector and the number of classes and $b$ is the bias vector.
	
	We used SVM as classifier in our experiments as SVM performs better than softmax classifier. SVM classifiers works on a margin based function while softmax classifier reduces the crossentropy function. 
	Multiclass SVM classifies data by locating the hyperplane
	at a position where all data points are classified correctly.
	Thus SVM determines the maximum margin among the data
	points of various classes~\cite{tang2013deep}. The more rigorous nature of
	classification of SVM makes it more suitable choice over softmax classifier.
		
	\begin{table}
		\centering
		\begin{tabular}{|c|c|}
			
			\hline 
			\textbf{Training Parameters} & \textbf{Values}   \\\hline 
			Momentum  &      0.9 \\\hline
			Initial Learn Rate  &      0.001 \\\hline
			Learn Rate Drop Factor  &      0.5 \\\hline
			Learn Rate Drop Period  &      10 \\\hline
			$L_2$ Regularization  &      0.004 \\\hline
			Max Epochs  &      70\\\hline
			MiniBatchSize  &   64 \\\hline		
			
		\end{tabular}
		\caption{Training Parameters for CNN}
		\label{tab:parameters for CNN}
	\end{table}

	\section{Experiments and Results}
	We experiment on three publicly available inertial datasets for human action recognition. These datasets are : Inertial component of UTD Multimodal
	Human Action Dataset (UTD-MHAD)~\cite{chen2015utd},"Heterogeneity Human Activity Recognition Dataset" (HHAR),~\cite{stisen2015smart} and the inertial component of Kinect V2 dataset~\cite{blog}. We used subject specific setting for experiments on both datasets by randomly splitting 80\% data into training and 20\% data into testing samples.
	 We ran the random split 20 times and report the average accuracy. We conduct our experiments on Matlab R2018b on a desktop computer with NVIDIA GTX-1070 GPU.
	 
	 \begin{figure*}[h]
	 	\centering
	 	\includegraphics[width=\linewidth]{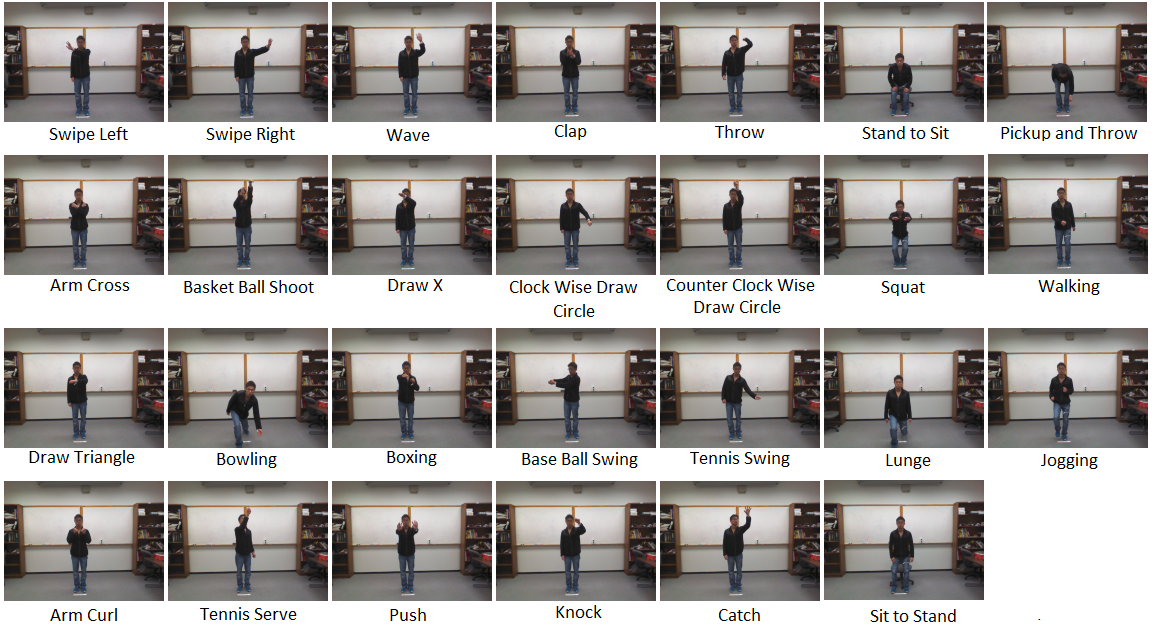}
	 	\caption{Sample actions from the UTD-MHAD data set}
	 	\label{fig:Samples}
	 \end{figure*}

	\subsection{UTD MHAD Dataset}
	
	This dataset consists of 27 different actions performed by 8 subjects with each subject repeating actions 4 times. These actions are shown in Fig.~\ref{fig:Samples}. 
	 We only used inertial component of the dataset and convert inertial time series data into signal images. We only obtained 1722 signal images which are very less to train CNN. Thus to increase the number of samples we used data augmentation technique described in~\cite{ahmad2018towards}. After augmentation we get 13776 samples of signal images.
	We used 11021 for training and 2755 for testing. We train
	CNN on signal images as shown in Fig.~\ref{fig:CNN Architecture for signal images} for 70 epochs. We achieve accuracy of 95.8\% using our proposed method.
	\begin{table}[h]
		\centering
		\begin{tabular}{|c|c|}
			
			\hline 
			\textbf{Previous Methods} & \textbf{Accuracy\%}  \\\hline 
		
			Chen et al.~\cite{chen2015real}      &      88.3 \\\hline
			Z.Ahmad et al.~\cite{ahmad2018towards}        &      93.7 \\\hline
			\textbf{Proposed Method}  & \textbf{95.8} \\
			\hline				
		\end{tabular}
		\caption{ Comparison of Accuracies of proposed method with previous methods on inertial component of UTD-MHAD dataset}
		\label{tab:comparisonTabe ON UTD-MHAD}
	\end{table}
	
	\subsection{Heterogeneity Human Activity Recognition Dataset}
	This dataset consists of six actions and is collected through nine users. All users followed a scripted set of activities while carrying eight smartphones (2 instances of LG Nexus 4, Samsung Galaxy
	S+ and Samsung Galaxy S3 and S3 mini) and four smart watches (2
	instances of LG G and Samsung Galaxy Gear). These smart phones and smart watches are worn by the participents at their waists and arms respectively. Each participant
	conducted five minutes of each activity, which ensured a near equal data
	distribution among activity classes (for each user and device). The six actions are : "Biking", "Sitting", "Standing", "Walking", "Stair Up" and "Stair down". We used 53760 samples for training and 13440 samples for testing. We did not perform data augmentation on this dataset and obtained an accuracy of 98.1\% using proposed method.
	
	\begin{table}[h]
		\centering
		\begin{tabular}{|c|c|}
			
			\hline 
			\textbf{Previous Methods} & \textbf{Accuracy\%}  \\\hline 
			
			Yao et al.~\cite{yao2017deepsense}    &      94.5 \\\hline
			
			\textbf{Proposed Method}  & \textbf{98.1} \\
			\hline				
		\end{tabular}
		\caption{ Comparison of Accuracies of proposed method with previous methods on HHAR dataset}
		\label{tab:comparisonTabe on HHAR}
	\end{table}
	
	\subsection{Kinect V2 Dataset}
	
	We use only inertial modality of KinectV2 action dataset. It contains 10 actions performed by six subjects with each subject repeating the action 5 times. The 10 actions are "right hand high wave", "right hand catch", "right hand high throw", "right hand draw X", "right hand draw tick", "right hand draw circle", "right hand horizontal wave", "right hand forward punch", "right hand hammer, and "hand clap".
	To increase the total number of samples, we perform data augmentation on the signal images obtained from the dataset. We select 3533 samples for training and 833
	samples for testing and obtained the accuracy of 98.3\%.
	
		\begin{table}[h]
		\centering
		\begin{tabular}{|c|c|}
			
			\hline 
			\textbf{Previous Methods} & \textbf{Accuracy\%}  \\\hline 
			
			Chen et al.~\cite{chen2016fusion}     &      96.7 \\\hline
			Z.Ahmad et al.~\cite{ahmad2018towards}        &      96.7 \\\hline
			\textbf{Proposed Method}  & \textbf{98.3} \\
			\hline				
		\end{tabular}
		\caption{ Comparison of Accuracies of proposed method with previous methods on inertial component of Kinect V2 dataset}
		\label{tab:comparisonTabe on Kinect V2}
	\end{table}

	\section{Conclusion}
	
	In this paper, we present a novel multidomain multimodal fusion framework that is capable of extracting discriminative and complementary features from the datasets. We made the input modality multidomain and multimodal by transforming spatial domain information into frequency and time-spectrum domain using Discrete fourier transform (DFT) and Gabor wavelet transform respectively. We extract features by employing CNN and then perform canonical correlation based fusion to get most informative set of features. We perform experiments on three inertial datasets and beat the previous state of art by considerable margin, thus demonstrating the usefulness of multidomain multimodal feature fusion. 
	
	\bibliographystyle{IEEEtran}

\end{document}